%% file: main.tex
\documentclass[10pt,twocolumn,letterpaper]{article}

\usepackage[final]{cvpr}     

\usepackage{multirow}
\definecolor{cvprblue}{rgb}{0.21,0.49,0.74}
\usepackage[pagebackref,breaklinks,colorlinks,allcolors=cvprblue]{hyperref}
\usepackage{float}

\title{Robotic Visual Instruction}
\author{%
Yanbang Li {$^{1}$\footnotemark[2]} ~~ Ziyang Gong {$^{2}$}~~ Haoyang Li {$^{3}$} ~~ Xiaoqi Huang {$^{4}$} ~~ Haolan Kang {$^{5}$}~~ \\ Guangping Bai {$^{6}$}~~ Xianzheng Ma {$^{6}$}~ \\
\normalsize
$^{1}$\ Imperial College London ~~$^{2}$\ Shanghai AI Laboratory~~ $^{3}$\ UC San Diego ~~ $^{4}$\ VIVO\\
\normalsize
$^{5}$\ South China University of Technology~~$^{6}$\ Independent Researcher
\normalsize
\footnotemark[2]{~~Corresponding Author}\\
\normalsize
\href{https://robotic-visual-instruction.github.io/}{\textcolor{pink}{https://robotic-visual-instruction.github.io/}}  
}

\begin{document}
\maketitle

\input{sec/0_abstract}    
\input{sec/1_introduction}

\input{sec/2_relatedwork}

\input{sec/3_method}

\input{sec/4_EXPC}
\input{sec/5_conclusion}

{
    \small
    \bibliographystyle{ieeenat_fullname}
    \bibliography{main}
}

\end{document}

%% file: sec/0_abstract.tex
\begin{abstract}
Recently, natural language has been the primary medium for human-robot interaction. 
However, its inherent lack of spatial precision introduces challenges for robotic task definition such as ambiguity and verbosity . Moreover, in some public settings where quiet is required, such as libraries or hospitals, verbal communication with robots is inappropriate.  To address these limitations, we introduce the\textbf{ Robotic Visual Instruction (RoVI)}, a novel paradigm to guide robotic tasks through an object-centric, hand-drawn symbolic representation. RoVI effectively encodes spatial-temporal information into human-interpretable visual instructions through 2D sketches, utilizing arrows, circles, colors, and numbers to direct 3D robotic manipulation.
To enable robots to understand RoVI better and generate precise actions based on RoVI, we present \textbf{Visual Instruction Embodied Workflow (VIEW)}, a pipeline formulated for RoVI-conditioned policies.
This approach leverages Vision-Language Models (VLMs) to interpret RoVI inputs, decode spatial and temporal constraints from 2D pixel space via keypoint extraction, and then transform them into executable 3D action sequences.
We additionally curate a specialized dataset of 15K instances to fine-tune small VLMs for edge deployment,
 enabling them to effectively learn RoVI capabilities. 
Our approach is rigorously validated across 11 novel tasks in both real and simulated environments, demonstrating significant generalization capability. Notably, VIEW achieves an 87.5\% success rate in real-world scenarios involving unseen tasks that feature multi-step actions, with disturbances, and trajectory-following requirements. \textcolor[rgb]{1, 0.42, 0.54}{https://robotic-visual-instruction.github.io/}

\end{abstract}

%% file: sec/1_introduction.tex
\begin{figure*}[t]
    \centering
    \includegraphics[width=1\linewidth]{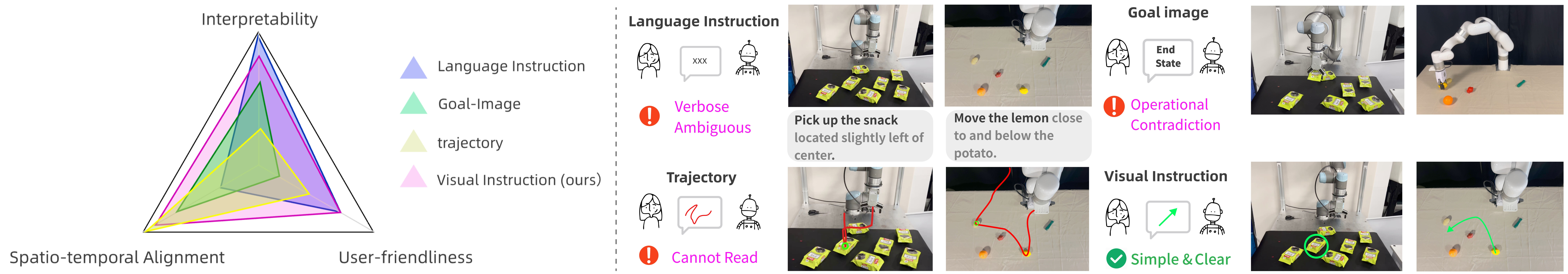}
    \caption{(Left) RoVI achieves an optimal balance of user-friendliness, interpretability, and spatiotemporal alignment. (Right) It shows examples and corresponding pros and cons of four types of human-robot interaction methods.}
    \label{fig:2}
\end{figure*}
\section{Introduction}
\label{sec:intro}

Natural language, is not always the optimal medium between humans and robots. Alternatively, sketching visual instructions convey more precise spatiotemporal information. 
Traditionally, communication between humans and robots relies on natural language, leveraging the advances in large language models (LLMs) to convert verbal or textual language instructions into executable actions for robots ~\cite{rt12022arxiv,rt22023arxiv,kim2024openvlaopensourcevisionlanguageactionmodel,lynch2022interactivelanguagetalkingrobots}.
While natural language is an intuitive and convenient medium for Human-Robot Interaction (HRI), it presents certain challenges.
Specifically, natural language has difficulty in describing spatial details such as the precise position, direction, or distance of objects~\cite{hassani2016visualizing, chang2024survey}. It is also prone to generating ambiguity and verbosity when expressing spatial requirements ~\cite{bonarini2020communication, wang2024large} shown in 
Figure \ref{fig:2}. 
Moreover, in certain public environments, such as libraries and hospitals, verbal communication may be inappropriate.

In contrast, visual modalities—such as goal images~\cite{sundaresan2024rt,inproceedings,Danielczuk_2019}, trajectories~\cite{gu2023rt,zhen20243d,xu2024flowcrossdomainmanipulationinterface}, and subgoal images~\cite{sundaresan2024rt,kang2024incorporatingtaskprogressknowledge}—offer a more direct and precise means of conveying spatio-temporal information. However, the practical application of such methods is not user-friendly shown in Figure \ref{fig:2}. The goal image requires the input of the end state of the robotic arm and the scene upon task completion, which contradicts the user's operational sequence. On the other hand, the trajectory represents the complete path of the end effector from the first to the last frame, posing challenges for users to imagine and draw the entire motion process of the robotic arm, which reduces the overall readability for users.

To address these limitations, we propose a novel communication paradigm: \textbf{Robotic Visual Instruction (RoVI)} shown in the left part of Figure~\ref{teaser}, \textit{which is a hand-drawn sketch instruction method, an object-centric representation that utilizes 2D symbolic language to command 3D embodiments.} 
This paradigm offers an intuitive, concise, and silent alternative to natural language instruction. Its basic primitives include arrows, circles, and various colors to represent different temporal sequences of actions, and numbers to label different embodiments for dual-arm systems. The arrows indicate the trajectory and direction, while the circles denote affordance location to identify target objects in a cluttered environment. Colors clearly convey the temporal sequence. 
By integrating these elements, RoVI compresses a temporal series of 3D coordinates into a human-understandable 2D visual language, thereby achieving an optimal balance of user-friendliness, interpretability, and spatiotemporal alignment, as shown in Figure~\ref{fig:2} left.

In order to better understand RoVI and use it to guide robotic manipulation, we introduce \textbf{ Visual Instruction Embodied Workflow (VIEW)}, a pipeline that transduces two-dimensional RoVI instructions into action sequences for robotic manipulation. 
VIEW facilitates the robotic interpretation of visual instructions, and translates them into hierarchical language responses and Python code functions via Vision-Language Models (VLMs). 
To decode temporal information from RoVI for high-level tasks, we decompose these tasks into multiple single-step subtasks based on color or numerical identifiers. 
Furthermore, we propose a keypoint module to extract keypoints from various RoVI components to serve as additional spatial and temporal constraints. 
Ultimately, our keypoint-conditioned policy directs the robot to execute the manipulation tasks, considering both spatial and temporal information from RoVI. 

Except for the framework, we develop a dataset of 15K training instances to enable models to learn RoVI capabilities through Parameter-Efficient Fine-Tuning (PEFT)~\cite{houlsby2019parameter, gong2024coda}.
Through the design outlined above, our approach performs well across diverse unseen tasks in both real-world and simulated environments, showing strong generalization and robustness.
Compared to language-conditioned policies, our method achieves superior performance in cluttered settings, multi-step operations, and trajectory-following tasks (see the right part of Figure~\ref{teaser}).

\begin{enumerate}
    \item We propose a novel human-robot interaction paradigm: RoVI. It employs hand-drawn symbolic representations as robotic instructions, conveying more precise spatial-temporal information within task definition. 
    \item We design a pipeline, VIEW (Visual Instruction Embodied Workflow), to enable RoVI-conditioned manipulation tasks. 
    \item We develop an open-source dataset to enable models to learn RoVI capabilities. The lightweight model trained by this dataset demonstrates that VLMs are able to learn this capability with minimal computational resources and simple fine-tuning. 
\end{enumerate}

%% file: sec/2_relatedwork.tex
\begin{figure*}[t]
    \centering
    \includegraphics[width=1\linewidth]{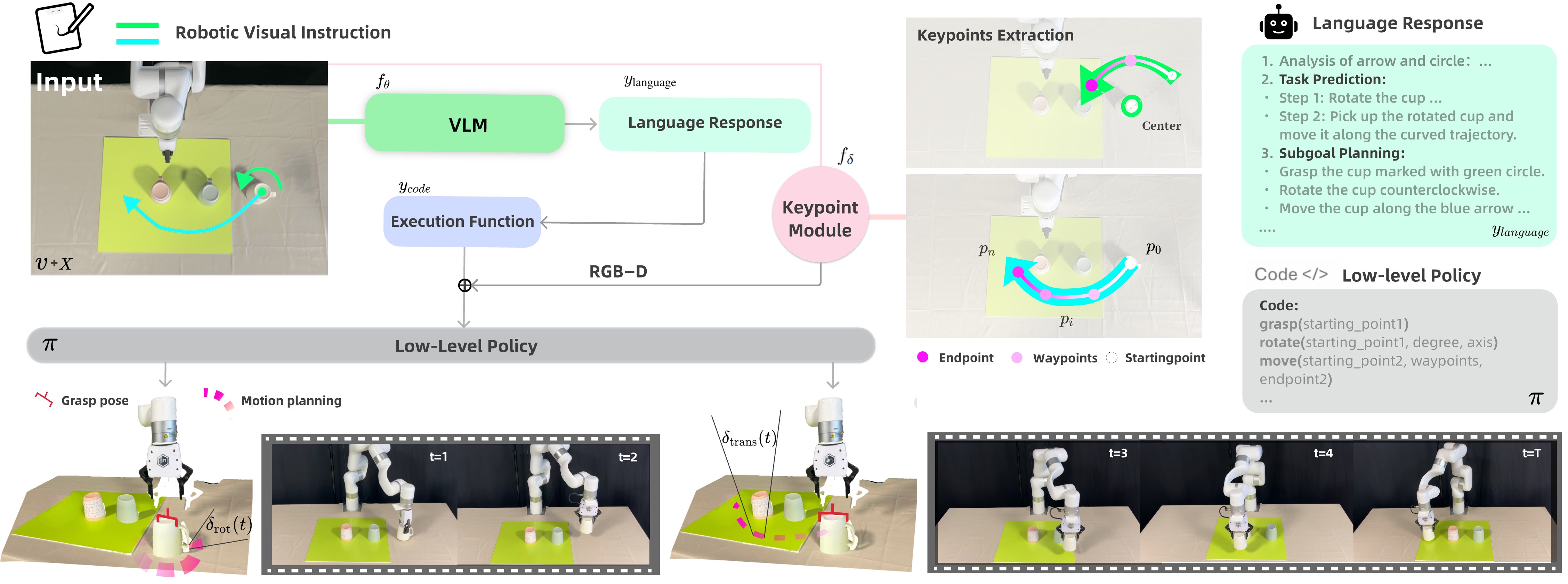}
    \caption{VIEW Architecture. This pipeline begins with a visual instruction drawn onto the initial observation. The VLM generates hierarchical sketch-to-action outputs, including task definition, detailed planning, and executable functions. The executable functions are then combined with keypoints extracted from the keypoint module and passed to a downstream low-level policy, which enables the robotic arm to execute each action step-by-step. This approach bridges hand-drawn visual instructions with precise robotic actions. }
    \label{fig:method}
\end{figure*}

\section{Related Work}
\label{sec:formatting}

\textbf{Human Robot Interaction.} Recent advancements in VLMs have made them a popular choice for language-conditioned policies
~\cite{brohan2022rt, brohan2023rt, kim2024openvlaopensourcevisionlanguageactionmodel,anonymousembodiedgpt,huang2023voxposer,ha2023scalingup}. Image-conditioned policies are also widely explored, such as goal-image policy ~\cite{sundaresan2024rt,kang2024incorporatingtaskprogressknowledge,Danielczuk_2019}, multimodal prompts~\cite{jiang2023vima}, and trajectory-based inputs~\cite{zhen20243d,gu2023rt,xu2024flowcrossdomainmanipulationinterface}. One common approach is goal-image conditioning, where a final goal image specifies the desired task's end state. Trajectory-based policy utilizes the full 2D or 3D trajectory of the end-effector as input. However, these input methods present significant challenges for users, as it is often difficult for users to provide such inputs directly in real-world applications.

\textbf{Visual Prompting for Robot.} Recent studies have explored the use of visual prompts as user input for tasks like Visual Question Answering (VQA)  \cite{yang2023dawn,achiam2023gpt,cai2024vip}. These models use symbolic forms of language, such as arrows, sketches, and numbers, to assist natural language in providing more accurate VQA. However, these approaches have primarily focused on generalized image-based question-answering tasks, and the domain of visual prompts in robotic manipulation tasks remains largely unexplored. In the context of robot control, some methods use model-generated visual prompts to guide trajectory selection for manipulation~\cite{liu2024moka,lee2024affordance,huang2024copa}.  Yet, they still rely on natural language as input. These methods do not resolve the issue with natural language instructions—specifically, the lack of spatial intent in task definitions provided by users. 

\textbf{Keypoint Constraints for Manipulation. }Recent studies \cite{huang2024rekep, shi2023waypointbasedimitationlearningrobotic,dipalo2024kat,jonnavittula2024view,Gao_2023} have achieved significant advancements in manipulation by leveraging key points to formulate spatiotemporal constraints. However, unlike prior approaches that extract keypoints from environmental objects and then filter them through VLMs reasoning \cite{huang2024rekep}, our method directly extracts key points from RoVI symbols (arrows and circles).


%% file: sec/3_method.tex
\section{Robotic Visual Instruction Design}
We present the paradigm design of RoVI, which consists of two visual primitives: an \textit{arrow} and a \textit{circle}. All simple or complex tasks are decomposed into three object-centric motions: moving from A to B (represented by an arrow), rotating an object (a circle indicating affordance with an arrow for rotation degree), and picking up/selecting (represented by a circle).

\textbf{Dissecting Arrow.} We use 2D \textit{arrows} to denote the trajectory and temporal sequence of robotic actions. An arrow is decomposed into three components: \textit{Tail} (Starting Point \( p_0 \)), \textit{Shaft} (Waypoints \( \{p_1, \dots, p_{n-1}\} \)), and \textit{Head} (Endpoint \( p_n \)). The starting point \( p_0 \) marks the grasp position on the object, and the endpoint \( p_n \) denotes the action's goal. Intermediate waypoints capture the movement path, forming an ordered set:
\begin{equation}
Arrow = \{ p_0, p_1, \dots, p_n \}, \quad p_i \in \mathbb{R}^2,
\end{equation}

where \( p_i \) are 2D coordinates extracted by a keypoint module.

\textbf{Dissecting Circle.} The \textit{circle} highlights key interaction areas on objects. The center point \( p_0 \in \mathbb{R}^2 \) represents the affordance center and is used for various tasks: as a grasping point, a pivot for rotation, or a pressure point for actions like pressing buttons.

\textbf{Drawing Setting.} 
RoVI is drawn directly using a stylus and drawing software on a tablet or PCs, with bright colors to ensure visibility across backgrounds: \textcolor[rgb]{0, 1, 0.37}{green} 
(RGB: 0, 255, 94) 
for first step of the manipulation task, \textcolor[rgb]{0, 1, 0.97}{blue} 
(RGB: 0, 255, 247) 
for the second step, and \textcolor[rgb]{1, 0.42, 0.54}{pink} 
(RGB: 255, 106, 138) 
for the third step. For more steps, extra color can be assigned flexibly. We designed two drawing styles: \textit{\textbf{Loose Style}} (casual, hand-drawn) and \textit{\textbf{Geometric Style}} (structured with geometric components for clearer interpretation by VLMs). We use a circle to signify affordances and replace the arrowhead with a standard triangle as depicted in Figure~\ref{design}. A comparison of their effectiveness is in Section~\ref{sec:abla}.

\begin{figure}
     \centering
     \includegraphics[width=1\linewidth]{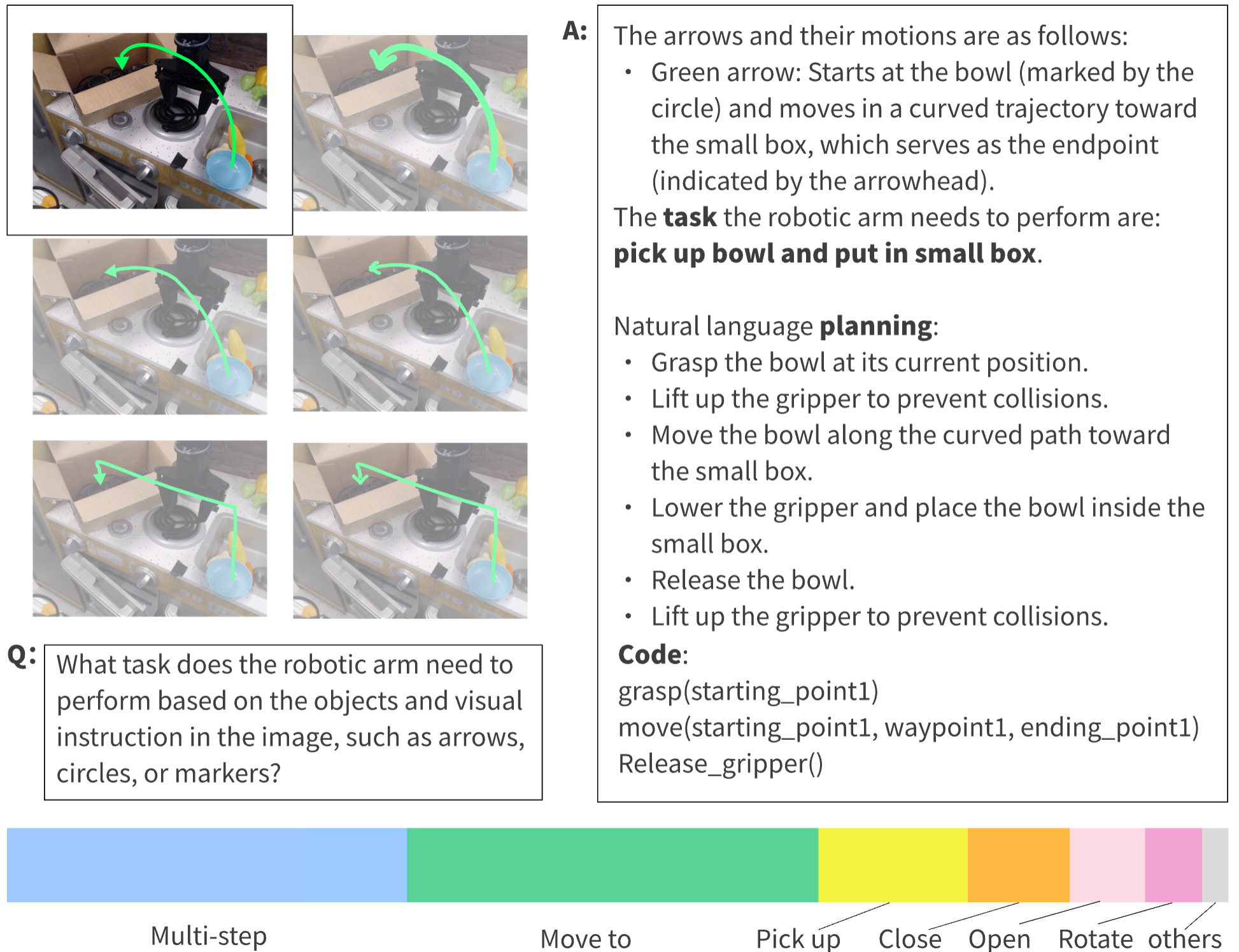}
     \caption{This is an example to demonstrate the RoVI Book dataset, adapted from the Open-X Embodiments dataset ~\cite{open_x_embodiment_rt_x_2023}. The bottom displays the proportion of each task type. }
     \label{fig:enter-label}
 \end{figure}

\section{RoVI Book dataset}

To enable VLMs to understand RoVI, we develop a dataset for RoVI-conditioned policy, termed \textbf{RoVI Book}. The dataset shown in Figure \ref{fig:enter-label} comprises 15K image-text question-answer pairs. It includes (1) images of initial task observations annotated with RoVI, (2) simple queries serving as default prompts, and (3) answers generated by GPT-4o~\cite{achiam2023gpt}, covering RoVI analysis, task names, fine-grained planning steps, and Python functions. The original tasks and images were selected from the Open-X Embodiment dataset~\cite{open_x_embodiment_rt_x_2023}. Our dataset covers 64\% single-step tasks and 36\% multi-step tasks, across five fundamental manipulation skills: \texttt{move an object}, \texttt{rotate an object}, \texttt{pick up}, \texttt{open drawers/cabinets}, and \texttt{close drawers/cabinets}. 
The answers are initially generated using GPT-4o~\cite{achiam2023gpt} and subsequently refined through semantic filtering based on human feedback. Each task retains its original semantic task name from the Open-X Embodiments~\cite{open_x_embodiment_rt_x_2023}, while we apply data augmentation to RoVI, introducing 3–8 visual variants, varying paths, drawing styles, and line thickness. Further details are provided in the appendix.

\section{Visual Instruction Embodied Workflow}
\subsection{Overview of Workflow}
\label{overview of pipeline}
The VIEW consists of three components: (1) A VLM $f_\theta$ for RoVI understanding and planning, (2) a keypoint module $ f_\delta$ for generating spatiotemporal constraints \cite{huang2024rekep}, and (3) a low-level policy $\pi$  for executing robot actions.

As shown in Figure \ref{fig:method}, the pipeline begins with VLMs that take as input the hand-drawn RoVI \( v \in \mathbb{R}^{H \times W \times 3} \), an initial observation image \( X \in \mathbb{R}^{H \times W \times 3} \), and a system-provided default prompt (further details on the default prompt can be found in the appendix). The VLMs then produce language response \( y_{\text{language}} \) and the execution function \( y_{\text{code}} \). Simultaneously, the keypoint module extracts keypoints from the RoVI to generate spatiotemporal constraints, including a starting point \( p_0 \), multiple waypoints \( p_i \), and an endpoint \( p_n \). Finally, based on the input \( y_{\text{code}} \) and the keypoint coordinates, the low-level policy executes the corresponding actions.

\subsection{VLMs for RoVI Understanding}

Given the VLMs' capabilities in visual perception, embedded world knowledge, and reasoning, we use the VLMs to interpret RoVI and translate it into a natural language response \( y_{\text{language}} \). The language response acts as a universal interface for human feedback, enabling verification of VLMs' comprehension and connecting it to downstream low-level policies. 
Compared with the end-to-end policies~\cite{rt12022arxiv, rt22023arxiv} directly output parameters in SE(3) action space, $y_{\text{language}}$ incorporates language-base action representations, which generalize more effectively across variable tasks and environments \cite{huang2023voxposer,rth2024arxiv,driess2023palme}.

The language response is generated by VLMs with a Chain-of-Thought (CoT) reasoning process. It includes coarse-grained task predictions, providing high-level task descriptions, and fine-grained planning with sub-goal sequences, breaking tasks into smaller steps.  Each sub-goal is subsequently converted into executable code functions \( y_{\text{code}} \), which define the necessary actions or skills for the robotic arm, such as \texttt{move()} or \texttt{grasp()}. These functions, combined with keypoint constraints, form a low-level policy for action implementation. A comprehensive example of the model output is provided in the appendix.
\begin{equation}
    y_{\text{language}}, y_{\text{code}} = f_\theta (v, X).
\end{equation}

\subsection{Keypoint Module}

To decode spatiotemporal information from RoVI, \( v \in \mathbb{R}^2 \) in pixel space, we first decompose multi-step tasks into single-step tasks based on color identifiers. The transition between single-step tasks is converted into motion between keypoints, 
specifically from the endpoint of the step \( j-1 \)  to the starting point of the step \( j \). 
Then, a trained keypoint module, \( f_\delta \),  provides keypoint constraints, which include sequences of end-effector coordinates and keypoints' semantic functionalities in manipulation such as starting points \( p_0 \in \mathbb{R}^2 \), waypoints \( p_i \in \mathbb{R}^2 \), and endpoints \( p_n \in \mathbb{R}^2 \).

We employ YOLOv8 \cite{Jocher_Ultralytics_YOLO_2023} as \( f_\delta \) and construct a dataset containing 2k images for its training (see details in the appendix). Compared to open-vocabulary object detection, our strategy simplifies the detection of all objects across different environments to identify components of the RoVI symbols, making it less susceptible to environmental variations or distractor objects (see Experiment Section~\ref{keypointEXP}).

\subsection{Keypoint-Conditioned Low-Level Policy}

We propose a keypoint-conditioned low-level policy that enables a robot to follow a sequence of target poses, defined as keypoints, for manipulation tasks. These keypoints \( p_i \in \mathbb{R}^2 \) are extracted from action arrows in an RGB image and mapped to 3D coordinates \( p'_i \in \mathbb{R}^3 \) using depth data from a RGB-D camera.

These  \( N \) keypoints are then mapped to a sequence of desired end-effector poses in SE(3) space, which is represented as  \( \{ \mathbf{e}_1, \mathbf{e}_2, \dots, \mathbf{e}_N \} \). The initial pose \( \mathbf{e}_0 \) is obtained using the grasp module \cite{fang2023anygrasp} based on \( p_0 \in \mathbb{R}^2 \).  
The series of poses form the action to be executed. We categorize actions into two types: \textit{translation} (e.g., move to, push, pull) and \textit{rotation} (e.g., flip, knock-down, adjust knob). 
At each time step \( t \), the robot performs:

\begin{enumerate}
    \item \textbf{State Observation}: Acquire the current end-effector pose \( \mathbf{e}_t \in \text{SE}(3) \) and target keypoint \( p'_i \in \mathbb{R}^3 \) from the RGB-D camera.
    
    \item \textbf{Cost Function Minimization}: \( \mathcal{L}_i(t) \): Minimize the cost function by moving towards \( p'_i \) leveraging motion planning and interpolation.

    \item \textbf{Keypoint Transition}: If \( \mathcal{L}_i(t) \leq \epsilon \), mark \( p'_i \) as reached and proceed to \( p'_{i+1} \). \( i \) accumulates until \( i = N \), then end the current action step.

\end{enumerate}

The goal at each time step \( t \) is to minimizes \( \mathcal{L}_i(t) \):
\begin{equation}
 \arg\min \, \mathcal{L}_i(t),
\end{equation}
\begin{equation}
\mathcal{L}_i(t) = \alpha_i \, \delta_{\text{trans}}(t) + (1 - \alpha_i) \, \delta_{\text{rot}}(t),
\end{equation}

where \( \alpha_i \) indicates the action type: \( \alpha_i = 1 \) for translation and \( \alpha_i = 0 \) for rotation.

\textbf{Translational Cost}: \( \delta_{\text{trans}}(t) = \left\| \mathbf{e}_t - \mathbf{e}_i \right\| \), where \( \mathbf{e}_t \) is the current end-effector pose and \( \mathbf{e}_i \) is the target pose, with \( \left\| \cdot \right\| \) denoting the Euclidean norm.

\textbf{Rotational Cost}: \( \delta_{\text{rot}}(t) = \left| \theta_t - \theta_i \right| \), where \( \theta_t \) is the current rotation angle, \( \theta_i \) is computed as:
\begin{equation}
\theta_i = \arccos\left( \frac{ (\mathbf{v}_i)^\top \mathbf{v}_{i+1} }{ \| \mathbf{v}_i \| \| \mathbf{v}_{i+1} \| } \right),
\end{equation}

with \( \mathbf{v}_i = p'_i - c \) and \( \mathbf{v}_{i+1} = p'_{i+1} - c \), where \( c \) is the rotation center.

%% file: sec/4_EXPC.tex
\input{Table/table2}
\begin{figure*}[h]
    \centering
    \includegraphics[width=1\linewidth]{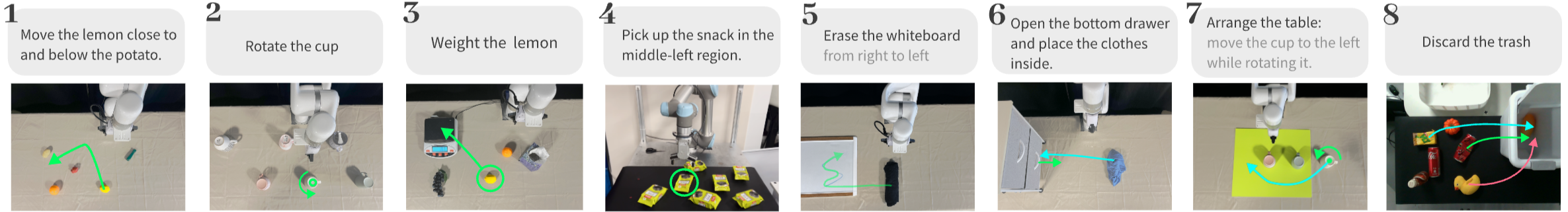}
    \caption{Robotic visual instruction is capable of generalizing to a variety of in-the-wild real-world situations, including multi-stage tasks that require precise spatial coordination, reasoning, and spatio-temporal dependencies, even in cluttered environments with disturbances.}
    \label{fig:real task}
\end{figure*}

\section{Experiment}

Our experiments aim to conduct in-depth research on the following questions:

\begin{enumerate}
    \item How does RoVI perform in generalizing over unseen environments and tasks in the real world and simulation? (section ~\ref{sec:6.1} and~\ref{sec:6.2}) 
    \item How well do current VLMs understand RoVI? (section ~\ref{sec:6.4}) 
    \item How do the components of RoVI and VIEW impact the overall performance of the whole pipeline? (section ~\ref{sec:abla})

\end{enumerate}

\textbf{Model Training.} We select GPT-4o~\cite{achiam2023gpt} and LLaVA-13B~\cite{liu2023LLaVA} as the VLMs in VIEW to control the robotic manipulation tasks. We also fine-tune the LLaVA-7B and 13B models \cite{liu2023LLaVA} using the LoRA~\cite{hu2022lora} on our RoVI Book dataset, with one training epoch and a learning rate of 2e-4. All experiments are conducted on an NVIDIA A40 GPU.


\textbf{Implement Procedure.} We train a YOLOv8 model \cite{Jocher_Ultralytics_YOLO_2023} to extract starting points, waypoints, and endpoints from hand-drawn instructions, providing keypoint constraints. These constraints are used to filter the grasp poses generated by AnyGrasp \cite{fang2023anygrasp} to obtain the closest one. The obtained 3D coordinates from RGB-D mapping and grasp poses are then input into VLM-generated Python functions for code-based low-level control.

\begin{figure}
    \centering
    \includegraphics[width=1\linewidth]{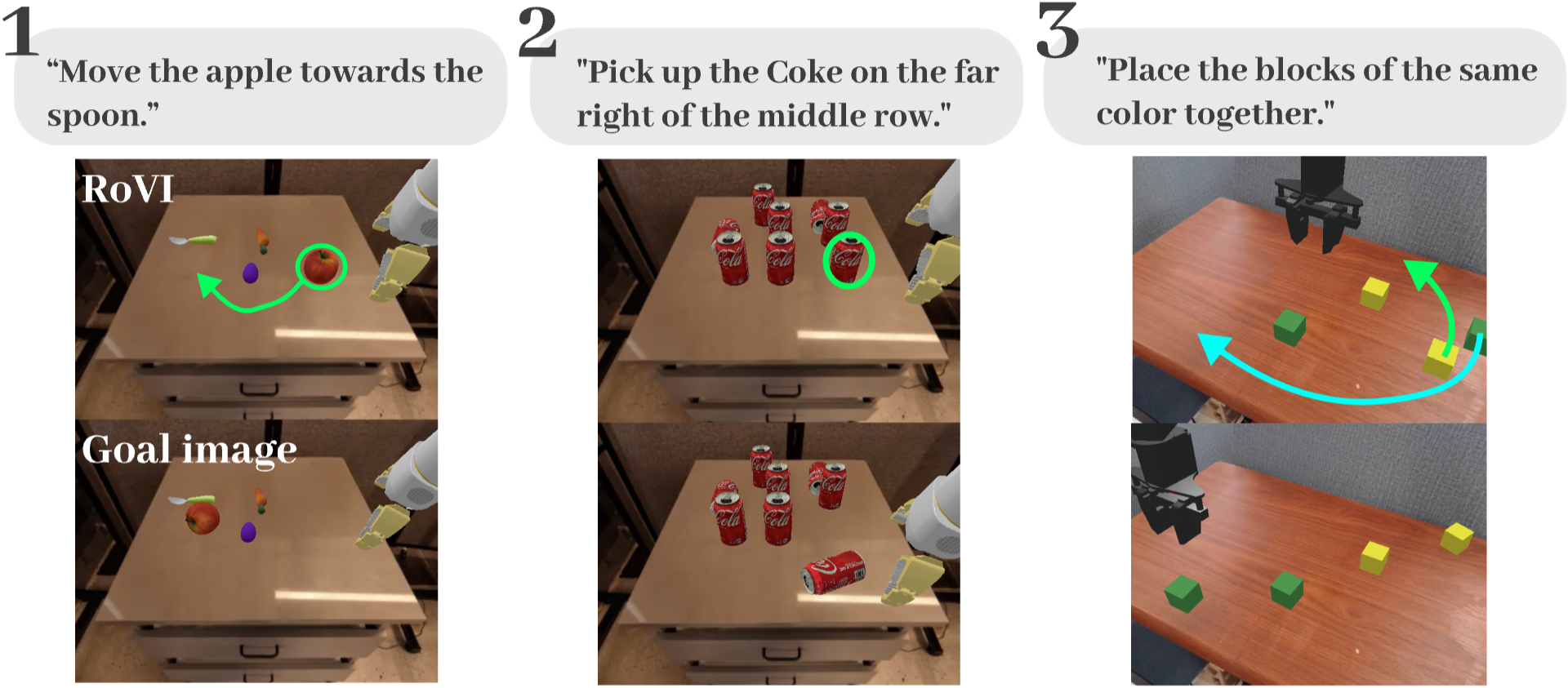}
    \caption{Experiment tasks in the SIMPLER ~\cite{li24simpler} environment for the comparative study of language instruction, goal image, and visual instruction. }
    \label{fig:simpler_env_task}
\end{figure}

\textbf{Manipulation Tasks.} We meticulously design 11 tasks: 8 in real environments and 3 in simulated settings shown in Figure \ref{fig:real task} and \ref{fig:simpler_env_task}. For our method, all tasks and environments are previously unseen, with new objects introduced. Our design includes 7 single-step tasks. Some involve cluttered environments with disturbances, such as `\texttt{select a desired object}' or `\texttt{move between objects}', requiring precise spatial alignment and trajectory-following abilities. Additionally, there are 4 multi-stage ( Task 6-8 in real environment, Task 3 in simulation) tasks to test further reasoning ability for spatio-temporal dependency.  

\input{Table/table1}

\subsection{Generalization to In-the-Wild Manipulation}
\label{sec:6.1}

\textbf{Real World Setting \& Baselines.} For real-world experiments, we use two robotic arms with two-finger grippers: UFACTORY \texttt{X-Arm 6} and \texttt{UR5}. Two calibrated RealSense D435 cameras are positioned for top-down and third-person views. Both robotic arms operated at a 20 Hz control frequency with an end-effector delta control mode. 

We compare our approach against two language-conditioned policy baselines, CoPa~\cite{huang2024copa} and VoxPoser~\cite{huang2023voxposer}, both leveraging a GPT model for low-level policy control. CoPa~\cite{huang2024copa} additionally utilizes Set-of-Mark (SoM)~\cite{yang2023setofmark} for object tagging as a visual prompt. To ensure a fair comparison, all methods used GPT-4o~\cite{achiam2023gpt} as the VLM.

\textbf{Evaluation Metrics for Action.} We report two metrics for assessing manipulation execution: \textbf{\textit{action success rate}}, measuring the percentage of tasks that meet defined goals, and \textbf{\textit{spatiotemporal alignment}}, evaluating the consistency of movement trajectories and the alignment of an object's final spatial state with semantic goals. A 6-point Likert scale is used for assessment (details in the appendix). Each task is evaluated over 10 trials.

\begin{figure}
    \centering
    \includegraphics[width=1\linewidth]{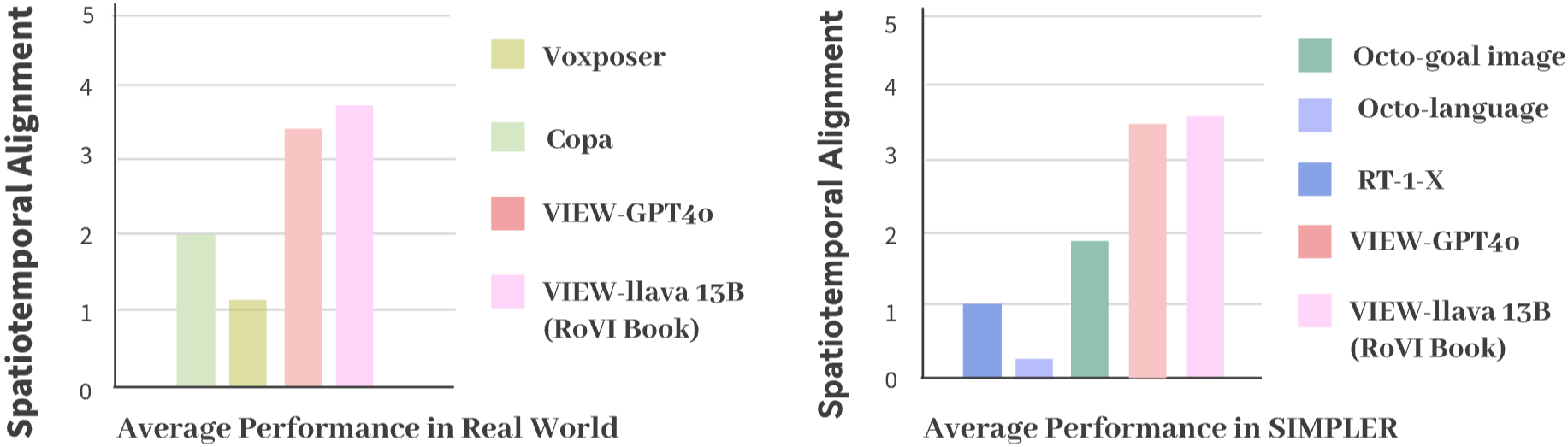}
    \caption{Performance comparison of average spatiotemporal alignment across all methods. See supplementary materials for detailed statistics.   }
    \label{fig:enter-label}
\end{figure}

\textbf{Results.} 
Table \ref{table2} shows that Voxposer~\cite{huang2023voxposer} and CoPa~\cite{huang2024copa} struggle with spatial precision tasks, such as `move the lemon close to and below the potato' and the `choose a snack' task with similar object disturbances. Both of these two methods also failed in Task 5, indicating the difficulty of the trajectory following. This is due to the inherent ambiguity of language-based instructions, which provide only object-level information, whereas RoVI enables pixel-level precision. In contrast, VIEW performs well on these tasks, as its keypoint module provides spatial constraints and waypoints. Unlike VoxPoser~\cite{huang2023voxposer} and CoPa~\cite{huang2024copa}, which use an open-vocabulary object detector, VIEW's keypoint module focuses on RoVI symbol parts, making it less susceptible to environmental variation or distractors. This enables VIEW's strong generalization and robustness in real-world manipulation tasks. Compared to other approaches that employ VLMs for temporal sequence reasoning in embodied planning, our method also achieves superior performance on long-horizon tasks (Task 6-8). By decomposing multi-step tasks into individual steps guided by color cues, we effectively reduce the complexity of temporal reasoning.

\subsection{Comparative Study in Simulation} \label{sec:6.2}

\textbf{Simulation Setting \& Baselines.} This section compares the manipulation performance of three instruction methods—language instruction, goal-image, and RoVI—in a simulated environment. We use SAPIEN~\cite{Xiang_2020_SAPIEN} as the simulator and SIMPLER~\cite{li24simpler} as the base environment. 

For the simulated experiments, we evaluate our approach against RT-1-X~\cite{rt12022arxiv} and Octo~\cite{octo_2023}, both of which are end-to-end, language-conditioned Vision-Language-Action (VLA) models trained on the Open X-Embodiment dataset~\cite{open_x_embodiment_rt_x_2023}. Octo~\cite{octo_2023} additionally supports goal-image input modalities. In our setup, we use the same robotic arms and background settings as in their training set and include new tasks in cluttered environments to test generalization.

\textbf{Quantitative Analysis.} These three tasks are performed in cluttered environments, where both language and goal-image inputs face significant challenges. Long-horizon tasks, in particular, are nearly impossible to accomplish under such conditions. However, our approach performs exceptionally well. These results indicate that end-to-end vision-language-action (VLA) models struggle with generalization to new tasks, while our method demonstrates robust generalization, with performance in simulation closely aligning with real-world outcomes.

\begin{figure}
    \centering
    \includegraphics[width=1\linewidth]{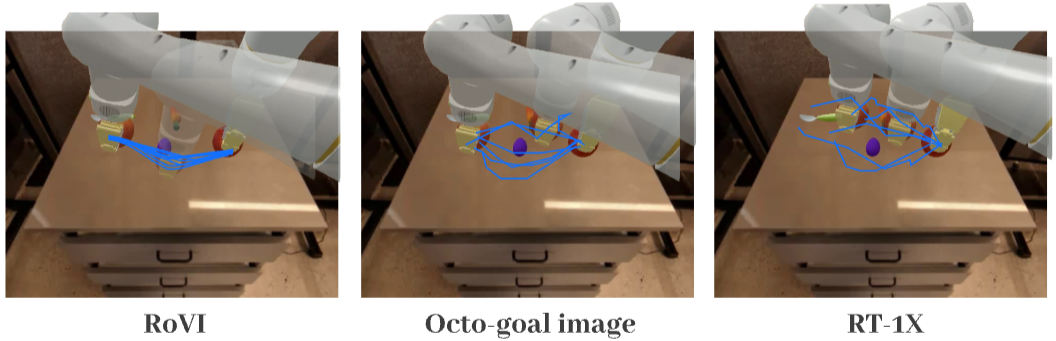}
    \caption{Visual comparison of trajectory between RoVI, natural language, and goal image policies. For each example, we sample six successful action trajectories from 50 trials and find that only RoVI's end state and path are more convergent and controllable.}
    \label{fig:trajectory}
\end{figure}

\textbf{Qualitative Study.}
To study the potential capability of RoVI, we delve into further qualitative comparison with nature language and goal-image conditioned policies. As shown in Figure \ref{fig:trajectory}, RoVI is the only instruction format that effectively conveys both path information and the end state. In contrast, the goal image policy performs well in terms of the end state but falls short in describing movement paths. For methods like RT-X~\cite{rt12022arxiv} and Octo~\cite{octo_2023}, the generated paths and end states lack consistency and exhibit limited spatial precision. In the evaluated examples, RoVI demonstrates a clear advantage in spatiotemporal alignment.

\begin{figure}
    \centering
    \includegraphics[width=1\linewidth]{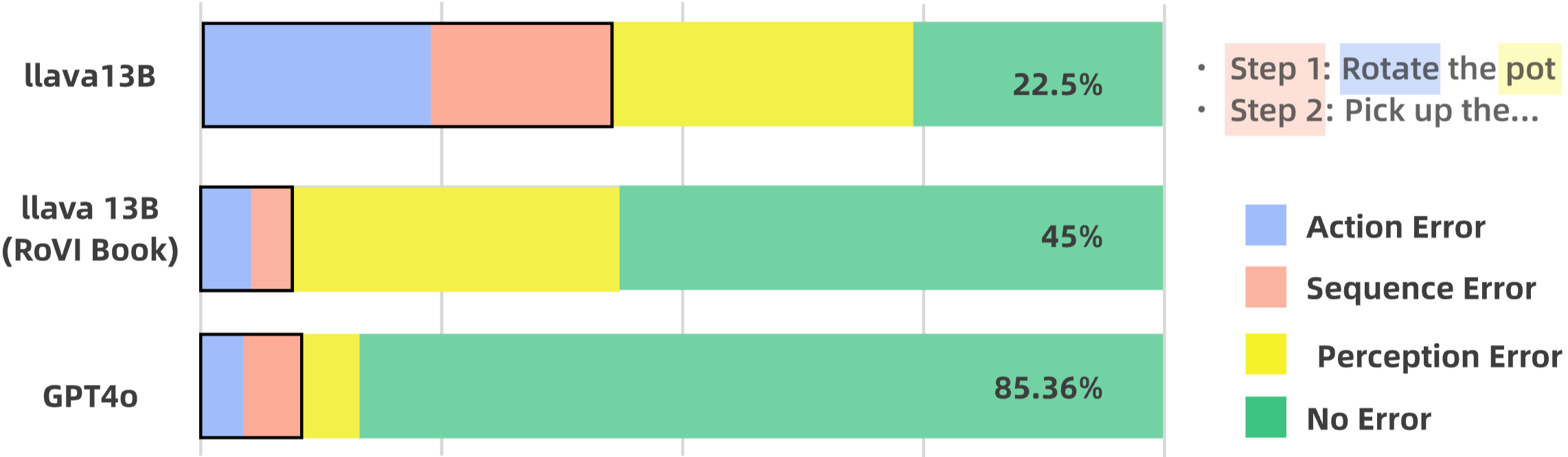}
    \caption{Error breakdown of language responses. Training with the RoVI book significantly reduces errors in action decisions and temporal sequences (highlighted in the black box).}
    \label{fig:errorbreakdown}
\end{figure}
\subsection{RoVI Comprehension by Modern VLMs.}
\label{sec:6.4}

We evaluate the capability of VLMs to extract semantic meaning from RoVI in novel tasks and environments, employing in-context learning and a zero-shot approach (see details in supplementary in-context learning). 

\textbf{Metrics.} We evaluate \textbf{\textit{`Task and Planning'}} success rates by assessing the accuracy of language responses using human feedback. This evaluation has two components: \textbf{\textit{`task'}}, measuring the VLMs' comprehension of task definitions based on RoVI and observations (e.g. `Open the bottom drawer, then place the clothes inside'); and \textbf{\textit{`planning'}}, evaluating the reasoning capability of VLMs to decompose complex RoVI tasks into sequential sub-goals. Each task is evaluated over 10 trials. We compare our trained model with diverse VLMs, including large-scale models: GPT-4o~\cite{achiam2023gpt}, Gemini-1.5 Pro~\cite{anil2023gemini}, Claude 3.5-Sonnet~\cite{TheC3}, as well as smaller models: InternLM-XComposer2-VL-7B~\cite{internlmxcomposer2}, LLaVA-HF/LLaVA-v1.6-Mistral-7B~\cite{liu2023LLaVA}, MiniGPT-4~\cite{zhu2023minigpt}, and VIP-LLaVA 7B~\cite{cai2024vip}.

\textbf{Results.} The Table \ref{table1} demonstrates that advanced large models (Gemini~\cite{anil2023gemini}, GPT-4o~\cite{achiam2023gpt}, Claude~\cite{TheC3}) exhibit a strong ability to understand RoVI-conditioned manipulation tasks through in-context learning, even without being trained on expert datasets. In contrast, models with fewer than 13 billion parameters fail to comprehend RoVI effectively. Combining both simulation and real-world performance, GPT-4o~\cite{achiam2023gpt} exhibits the best overall results. Furthermore, advanced large models generalize better in terms of RoVI comprehension compared to smaller models trained on the RoVI Book dataset, such as LLaVA-13B~\cite{liu2023LLaVA}. However, as the number of steps in the task increases, the large models' comprehension accuracy decreases. In contrast, LLaVA-13B~\cite{liu2023LLaVA}, trained on the RoVI Book dataset, performs well on long-sequence task 8, indicating that the RoVI Book dataset is effective for learning multi-step tasks under RoVI conditions.

\textbf{Error Breakdown.} It is worth noting that LLaVA-13B~\cite{liu2023LLaVA} (trained on the RoVI Book) shows a low success rate in task and planning predictions but performs exceptionally well in action execution. In conjunction with Figure \ref{fig:errorbreakdown}, we can conclude that the execution function maps action and sequence errors, making it unaffected by perception errors. After training on the RoVI Book, errors related to the execution function were significantly reduced.

\subsection{Ablations Study }
\label{sec:abla}

\begin{figure}[t]
    \centering
    \includegraphics[width=1\linewidth]{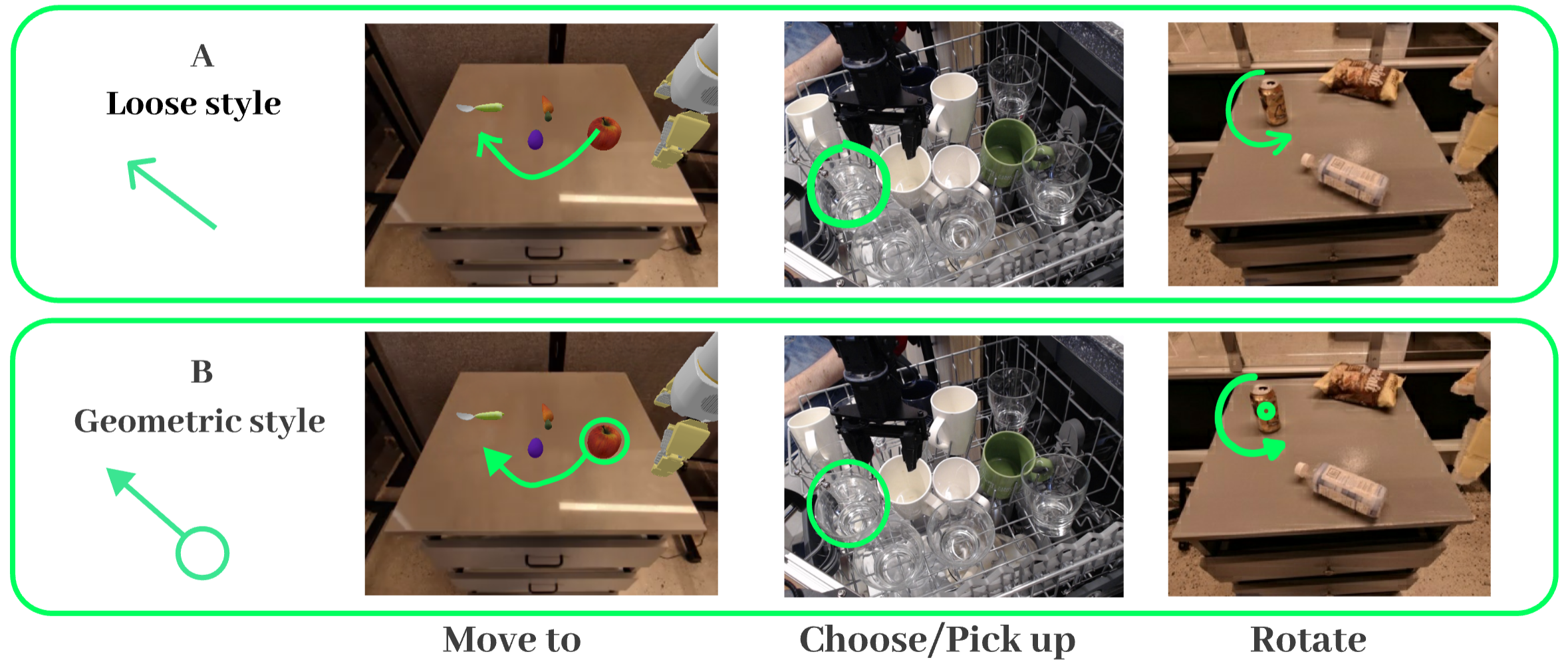}
    \caption{Showcase of two drawing styles in modified Open X-Embodiment dataset~\cite{open_x_embodiment_rt_x_2023}.}
    \label{design}
    
\end{figure}

\textbf{Drawing. }
Analogous to how language prompts often require `prompt engineering', free-form drawing can exhibit significant variability. And hand-drawn instruction raises another question: how can we optimize the drawing style to enhance model comprehension? In this section, we classify the drawing styles into two distinct categories for comparison to investigate their impact on VLMs' reasoning performance. The corresponding visualization and experiments are shown in Figure \ref{design} and Table \ref{total ablation}. Our findings indicate that the more structured geometric style yielded superior comprehension. 
Further experimental details are attached in the supplementary material.

\input{Table/table4}
\input{Table/table5}

\textbf{Keypoint Module.} \label{keypointEXP} We evaluate the proposed keypoint module, a trained YOLOv8 model~\cite{Jocher_Ultralytics_YOLO_2023}, for spatial constraint generation across four different RoVI tasks. We compare it against three popular open-vocabulary detection models~\cite{liu2023grounding,minderer2022simple,minderer2024scaling}, using two strategies: (1) manually inputting the target’s semantic information as the text prompt, and (2) identifying and localizing arrow components (arrowhead and tail).  Two primary metrics are used for evaluation: Euclidean distance error (measured in pixels) to assess precision, and Mean Average Precision (mAP) at a 50-pixel threshold to measure accuracy. Results in Table \ref{keypoint ablation} indicate that, despite its smaller parameter size, the keypoint module achieves more efficient task-relevant keypoint extraction directly from pixel space compared to transformer-based open-vocabulary detection models. Additional limitations and details can be found in the supplementary material.

%% file: Table/table2.tex
\begin{table*}[]
\centering
{\fontsize{8pt}{12pt}\selectfont
\resizebox{1\linewidth}{!}{
\begin{tabular}{cccccccccc|ccccc}
    \toprule
    \multirow{2}{*}{Robotic Baseline} & \multicolumn{8}{c}{Real World} &\multirow{2}{*}{Average} & \multirow{2}{*}{Robotic Baseline} & \multicolumn{3}{c}{Simulator} &\multirow{2}{*}{Average}\\
    \cmidrule{2-9} \cmidrule{12-14}
    &1 & 2 & 3 & 4 & 5 & 6 & 7 & 8 & & & 1 &2 &3&  \\
    \midrule
    \multirow{1}{*}{Voxposer~\cite{huang2023voxposer}} & 30 & 80 & 80  & 10 & 0 & 30 & 30 & 20 & 43.8 & RT-1-X~\cite{rt12022arxiv}& \textbf{40}& \textbf{20} &0& \textbf{20}\\
    \multirow{1}{*}{CoPa~\cite{huang2024copa}} & 40  & 90 & 80 & 60 & 0 & 40 & 20 & 30 & 45& Octo-goal-image~\cite{octo_2023} & 10 &30& 0& 13.3\\

    \multirow{1}{*}{VIEW-GPT4o~\cite{achiam2023gpt}} & 80 & \textbf{100} & 90 & 90 &  60 & \textbf{70}& \textbf{90} & 80 & 82.5 & Octo-language~\cite{octo_2023} & 10&0&0&3\\
    \multirow{1}{*}{VIEW-LLaVA-13B \cite{liu2023LLaVA} (RoVI Book)} & \textbf{90} & 90 & \textbf{100} & \textbf{100} & \textbf{70} & \textbf{70} & \textbf{90} & \textbf{90} & \textbf{87.5} &VIEW$^*$ & \textbf{70}&\textbf{60}& \textbf{100}&\textbf{76.6}\\
    \bottomrule
\end{tabular}}}
\caption{Success Rate in unseen environments and unseen tasks. The numbers correspond to tasks in Figure ~\ref{fig:real task} and Figure~\ref{fig:simpler_env_task}. VIEW$^*$ denotes both VIEW-GPT4o and VIEW-LLaVA 13B (RoVI Book), as their test results are identical.\textbf{ Bold} score means the best result.} 
\label{table2}
\end{table*}

%% file: Table/table1.tex
\begin{table*}[]
\centering
{\fontsize{5.6pt}{8pt}\selectfont
\resizebox{1\linewidth}{!}{
\begin{tabular}{cccccccccc|cccc}
    \toprule
    \multirow{3}{*}{VLMs w/ RoVI}  & \multicolumn{9}{c}{Real World}& \multicolumn{4}{c}{Simulator}\\
    \cmidrule{2-10} \cmidrule{11-14}
    & 1 & 2 & 3 & 4 & 5 & 6 & 7 & 8 & Average & 1 &  2 &3 & Average \\
    \midrule
    \multirow{1}{*}{Small Models}   & 0 & 0 & 0 & 0 & 0 & 0 & 0 & 0 & 0& 0 & 0 & 0& 0 \\
    \multirow{1}{*}{Claude 3.5-Sonnet~\cite{TheC3}}   & \textbf{100} & 95 & 0 & \textbf{100} & \textbf{90} & 55 & 50 & 67 & 70 & 30 & \textbf{100} & 50& 60\\
    \multirow{1}{*}{Gemini-1.5 Pro~\cite{anil2023gemini}}   & 10 & \textbf{100} & \textbf{100} & \textbf{100} & 20 & \textbf{95} & 60 & 57 & 68 & 0 & \textbf{100} & 0 & 33 \\
    \multirow{1}{*}{GPT-4o~\cite{achiam2023gpt}}   & \textbf{100} & \textbf{100} & \textbf{100} & 40 & 60& 90 & \textbf{100} & 55 & \textbf{81} & \textbf{100} & \textbf{100} & \textbf{90}& \textbf{97}\\
    \multirow{1}{*}{LLaVA-13B \cite{liu2023LLaVA} (RoVI Book)}   & 9 & 45 & 0 & 82 & 0 & 75 & 14 & \textbf{82} & 38 & 36 & 82 & 73& 64\\
    \bottomrule
\end{tabular}}}
\caption{Task and Planning evaluation in language response. It showcases the capability of existing VLMs to comprehend RoVI.  The numbers correspond to tasks in Figure \ref{fig:real task} and Figure \ref{fig:simpler_env_task}.}
\label{table1}
\end{table*}

%% file: Table/table4.tex
\begin{table}[]
\centering
\scriptsize

\resizebox{\linewidth}{!}{
\begin{tabular}{ccc|cc|cc|cc}
\toprule
\multirow{5}{*}{Model} & \multicolumn{8}{c}{Task Prediction $+$ Subgoal Planning} \\
 \cmidrule(lr){2-9}
 &\multicolumn{2}{c}{Move} &\multicolumn{2}{c}{Pick up / Choose} & \multicolumn{2}{c}{Rotate} &\multicolumn{2}{c}{Average}
\\
\cmidrule(lr){2-3} \cmidrule(lr){4-5} \cmidrule(lr){6-7} \cmidrule(lr){8-9} 
 & L & G & L & G & L & G & L & G \\
 \midrule
 GPT-4o~\cite{achiam2023gpt} &0.6& \textbf{1.0}& \textbf{0.9}& \textbf{0.9}&0.2& \textbf{1.0}& 0.57& \textbf{0.97}\\
 Gemini 1.5 pro~\cite{anil2023gemini}& \textbf{0.1}& 0.0& \textbf{1.0} & \textbf{1.0} & \textbf{1.0}& 0.6&\textbf{0.7}& 0.53\\
 Claude 3.5 sonnet~\cite{TheC3} & \textbf{1.0}& 0.8& \textbf{1.0}& \textbf{1.0}& \textbf{0.9}& \textbf{0.9}& \textbf{0.97}& 0.9\\
Total Average & 0.57 & \textbf{0.6}& \textbf{0.97} &\textbf{0.97}& 0.7& \textbf{0.83}& 0.74& \textbf{0.8}\\
\bottomrule
\end{tabular}}
\caption{Comparison of drawing styles in modified
Open X-Embodiment. `L' and `G' denote Loose style and Geometric style respectively. On average,  the more structured geometric style offers VLMs better task comprehension ability.}
\label{total ablation}
\end{table}

%% file: Table/table5.tex
\begin{table}[htbp]
\centering

\resizebox{\linewidth}{!}{
\begin{tabular}{llcccc}
\toprule
\textbf{Task} & \textbf{Metric} & \textbf{GDINO~\cite{liu2023grounding}} & \textbf{OWL-ViT~\cite{minderer2022simple}} & \textbf{OWL-V2~\cite{minderer2024scaling}} & \textbf{YOLOv8~\cite{Jocher_Ultralytics_YOLO_2023}}\\
\midrule
\multirow{2}{*}{ 1} &  MD & 482.71 $\pm$ 0.00 & N/A & 114.18 $\pm$ 92.65 & \textbf{6.83 $\pm$ 0.00} \\
& mAP & 0.00 $\pm$ 0.00 & N/A & 0.33 $\pm$ 0.47 & \textbf{1.00 $\pm$ 0.00} \\
\midrule
\multirow{2}{*}{ 2} &  MD & 507.35 $\pm$ 183.27 & N/A & 52.47 $\pm$ 61.44 & \textbf{19.45 $\pm$ 8.92} \\
&   mAP & 0.00 $\pm$ 0.00 & N/A & 0.57 $\pm$ 0.49 & \textbf{1.00 $\pm$ 0.00} \\
\midrule
\multirow{2}{*}{ 3} &  MD & 510.33 $\pm$ 183.25 & 153.92 $\pm$ 0.00 & 131.03 $\pm$ 33.43 &\textbf{ 13.27 $\pm$ 5.81} \\
&   mAP & 0.00 $\pm$ 0.00 & 0.00 $\pm$ 0.00 & 0.00 $\pm$ 0.00 & \textbf{1.00 $\pm$ 0.00} \\
\midrule
\multirow{2}{*}{ 4} &  MD & 751.44 $\pm$ 196.85 & 57.51 $\pm$ 89.85 & 63.28 $\pm$ 80.47 & \textbf{11.64 $\pm$ 4.73} \\
&   mAP & 0.00 $\pm$ 0.00 & 0.67 $\pm$ 0.47 & 0.64 $\pm$ 0.48 & \textbf{1.00 $\pm$ 0.00} \\
\bottomrule

\end{tabular}}
\caption{Ablation of the proposed keypoint module. The tested tasks and RoVI are shown in the supplementary material. MD represents mean distance. }
\label{keypoint ablation}
\end{table}

%% file: sec/5_conclusion.tex
\section{Conclusion and Future works}
In this paper, we propose \textbf{Robotic Visual Instruction} \textbf{(RoVI)}, a user-friendly and spatially precise alternative to natural language for guiding robotic tasks. To implement RoVI, we develop a pipeline, \textbf{Visual Instruction Embodied Workflow} \textbf{(VIEW)}, which demonstrates strong generalization and robustness across cluttered environments and long-horizon tasks. Additionally, we meticulously create a dataset to fine-tune VLMs for a better understanding of RoVI and potential future edge device deployment. 
Ablation studies also reveal the factors influencing the performance of RoVI-conditioned policies, including RoVI comprehension, drawing strategies, and grounding methods. 


\textbf{Future works.} Future research will focus on scaling up the RoVI Book dataset and collecting a wider variety of free-form drawn instruction. This expansion aims to equip the model with a broader understanding of the general principles by which humans employ visual symbols to convey dynamic movements. 
On the other hand, we can more efficiently train a smaller model like 7B. This will facilitate the deployment of edge devices within our robotic system.